\documentclass[conference]{IEEEtran}
\IEEEoverridecommandlockouts
\usepackage{cite}
\usepackage{amsmath,amssymb,amsfonts}
\usepackage{algorithmic}
\usepackage{graphicx}
\usepackage{url}
\usepackage{textcomp}
\usepackage{xcolor}
\usepackage[ruled,vlined,linesnumbered]{algorithm2e}
\usepackage{amsmath}
\usepackage{amsfonts}
\usepackage{bm}
\usepackage{booktabs}
\usepackage{graphicx}
\def\BibTeX{{\rm B\kern-.05em{\sc i\kern-.025em b}\kern-.08em
    T\kern-.1667em\lower.7ex\hbox{E}\kern-.125emX}}
\begin{document}

\title{ES-AHD: An Evolution Strategy Framework for Automatic Heuristic Design
\thanks{}
}

\author{ \IEEEauthorblockN{ Yutao Lai\IEEEauthorrefmark{1}, Kezhao Lai\IEEEauthorrefmark{1}, Hai-Lin Liu\IEEEauthorrefmark{1}, Yuping Wang\IEEEauthorrefmark{2}, and Ping Guo\IEEEauthorrefmark{3} } \IEEEauthorblockA{ \IEEEauthorrefmark{1} \textit{School of Mathematics and Statistics}\\ \textit{Guangdong University of Technology}\\ Guangzhou, China\\ hlliu@gdut.edu.cn } \IEEEauthorblockA{ \IEEEauthorrefmark{2} \textit{School of Computer Science and Technology}\\ \textit{Xidian University}\\ Xi'an, China } \IEEEauthorblockA{ \IEEEauthorrefmark{3} \textit{School of Systems Science}\\ \textit{Beijing Normal University}\\ Beijing, China } }

\maketitle

\begin{abstract}
In this paper, we introduce ES-AHD, a novel framework that fundamentally integrates Evolution Strategy (ES) into Large Language Model (LLM)-driven Automatic Heuristic Design (AHD). Existing evolutionary approaches predominantly rely on random, individual-level mutation, leading to blind search and an imbalance between exploration and exploitation. To address these issues, ES-AHD introduces two core mechanisms. First, Semantic Recombination via LLMs discards traditional point-to-point reproduction. By leveraging the LLM's contextual reasoning to explicitly extract core insights from top-performing individuals, the algorithm establishes a promising semantic search direction. This transforms random code mutation into targeted, center-guided sampling inspired by ES. Second, Stochastic Covariance Adaptation via Temperature Sampling dynamically addresses the exploration-exploitation dilemma. By mapping the covariance matrix in ES to the LLM's sampling temperature, the framework employs a stochastic random walk mechanism with momentum. This approach primarily shrinks the search radius for micro-level code refinement, while retaining the critical ability to occasionally sample higher temperatures to escape semantic local optima. Ultimately, ES-AHD provides a highly directional, robust, and efficient search paradigm, significantly accelerating the generation of high-quality heuristic algorithms. The source code is available at: \url{https://github.com/Mriya0306/ES-AHD}.

\end{abstract}


\section{Introduction}
When tackling complex combinatorial optimization problems~\cite{peres2021combinatorial} in the real world, such as vehicle routing~\cite{el2025metaheuristic}, bin packing~\cite{sun2025two}, and scheduling~\cite{zhang2025literature}, exact algorithms often fail to find optimal solutions within polynomial time due to the NP-hard nature of these problems. Consequently, heuristic algorithms have become the mainstream choice for practical applications. Traditionally, the development of high-performance heuristics has heavily relied on manual design by domain experts. Experts must leverage profound mathematical knowledge, extensive trial-and-error, and deep insights into specific problem structures to craft strategies like greedy algorithms, local search, or meta-heuristic frameworks. However, this manual design paradigm is not only time-consuming and labor-intensive but also highly constrained by inherent human cognitive biases, making it exceedingly difficult to generalize hand-crafted algorithms across different problem domains. To overcome the bottlenecks of manual design, researchers introduced Automatic Heuristic Design (AHD). Traditional AHD methods, primarily represented by Genetic Programming (GP)~\cite{zhang2022multitask}, attempt to automate the design process by using basic operators and rules as building blocks, evolving new heuristic rules within a predefined search space. Although traditional AHD achieved partial automation, it faces fatal limitations: bounded by abstract syntax trees or low-level instruction sets, its expressiveness is severely restricted. The resulting search space is vast yet sparse, making it nearly impossible to generate high-performance heuristic algorithms with complex logic (e.g., advanced data structures, loops, and conditional branches) that are also human-readable. 

Recently, the unprecedented leap in the code generation and logical reasoning capabilities of Large Language Models (LLMs) has sparked a paradigm shift across multiple computational domains simultaneously. Researchers have actively leveraged LLMs to automate general software engineering tasks~\cite{jin2024llms,he2025llm,xia2024agentless}, navigate the complex structural spaces of Neural Architecture Search (NAS)~\cite{lai2026llmenas,cai2025seki,zhou2025design,RZ_NAS}, and design intelligent operators for evolutionary computation~\cite{liao2025llm4eo,huang2025autonomous}. In parallel with these widespread advancements, LLMs have also revolutionized the Automatic Heuristic Design (AHD) domain~\cite{liu2024evolution}. Pioneering works, such as FunSearch~\cite{romera2024mathematical} and Evolution of Heuristics~\cite{liu2024evolution} (EoH), have successfully integrated LLMs into evolutionary computation frameworks to directly evolve heuristics at the code level. These LLM-based AHD methods utilize the rich prior knowledge of LLMs as "crossover" and "mutation" operators, enabling the generation of logically complex and highly innovative algorithms. However, existing LLM-based AHD approaches are largely confined to the traditional Genetic Algorithm (GA) paradigm. They predominantly rely on point-to-point, random crossover, or mutation among a few parent code snippets. This individual-level micro-operation lacks the macro-level extraction of globally optimal strategies, leading to a highly blind search process. Furthermore, under fixed sampling temperature parameters, LLMs easily fall into hallucination loops where they repeatedly generate similarly flawed code, causing a severe imbalance between exploration and exploitation. To address the blind search and convergence imbalance inherent in current LLM-based heuristic evolution, we propose a novel framework: Evolution Strategy for Automatic Heuristic Design (ES-AHD). Breaking away from traditional individual-level crossover and mutation, we introduce the core philosophy of the evolution strategy~\cite{beyer2002evolution} into a high-dimensional semantic space. Specifically, the main contributions of this paper are twofold: 

 1) Semantic Recombination via LLMs: We discard blind, point-to-point mutation. By leveraging the contextual reflection capabilities of LLMs, the algorithm explicitly extracts a "Core Insight" from the Top-K performing candidate individuals. This insight serves as a promising semantic search direction, effectively upgrading random code mutation into center-guided, directional sampling. 

 2) Stochastic Covariance Adaptation via Temperature Sampling: We map the global search variance in Evolution Strategies to the LLM's sampling temperature. Rather than relying on a deterministic monotonic decay, we introduce a stochastic random walk mechanism modulated by a constant step-size. This formulation maintains momentum from previous generations while introducing Gaussian noise. Consequently, the framework achieves a robust balance: it smoothly shrinks the search radius for micro-level code refinement as iterations progress, yet crucially retains the ability to occasionally sample higher temperatures to escape semantic local optima.

 \section{Related work}
 
 The integration of LLMs into AHD has fundamentally transformed how optimization algorithms are generated. The pioneering work, FunSearch~\cite{romera2024mathematical}, demonstrated that LLMs could act as powerful evolutionary operators within an island-based genetic algorithm framework to discover novel programmatic heuristics. By pairing a creative LLM code generator with a rigorous automated evaluator, FunSearch successfully solved complex combinatorial problems that were difficult to tackle with traditional methods. To overcome the inefficiency of raw code mutation, EoH~\cite{liu2024evolution} evolves both natural language "thoughts" and executable "codes," leveraging the LLM's semantic understanding to translate conceptual ideas into high-performance heuristics.

As the field matured, researchers began addressing the limitations of single-objective optimization and blind search processes. Recognizing that practical applications require balancing multiple criteria (such as performance versus computational efficiency), the Multi-objective Evolution of Heuristic (MEoH) method~\cite{yao2025multi} was proposed as the LLM-driven multi-objective search framework. MEoH utilizes an LLM generator coupled with a novel dominance-dissimilarity population management mechanism to produce a diverse, non-dominated set of trade-off heuristics. Concurrently, to mitigate the blindness of traditional evolutionary operators, ReEvo~\cite{ye2024reevo} introduced a hyper-heuristic approach. ReEvo significantly improves sample efficiency by utilizing LLM "reflections" to analyze past successes and failures, providing verbal gradients that guide the targeted rewriting and refinement of heuristics within the search space.

Recent advancements have further expanded the scope of AHD from single-algorithm generation to holistic set design and systematic global exploration. Because a single heuristic often struggles to generalize across varying problem distributions, the Evolution of Heuristic Set (EoH-S)~\cite{EOH_S} framework shifted the objective toward Automated Heuristic Set Design. By employing complementary-aware memetic search and population management, EoH-S generates a small set of mutually complementary heuristics, ensuring robust performance across diverse problem instances. To address the issue of population-based methods prematurely discarding temporarily underperforming solutions and falling into local optima, MCTS-AHD~\cite{MCTS_AHD} integrated Monte Carlo Tree Search into the LLM-driven evolutionary process. By organizing all generated heuristics in a search tree and applying an exploration-decay technique, MCTS-AHD systematically balances exploration and exploitation, successfully allowing latent, underperforming heuristics the chance to evolve into state-of-the-art solutions. However, traditional MCTS evaluates and selects heuristics strictly at the individual node level, which remains highly inefficient when navigating the massive and sparse search spaces generated by LLMs. To overcome this sampling bottleneck, recent research introduced Clade-AHD~\cite{lai2026beyond}, a framework that advances selection beyond single nodes to broader evolutionary branches (clades). By evaluating joint values and allocating computational resources at the macro-clade level, this approach significantly enhances sampling efficiency in large-scale heuristic search, allowing the algorithm to more accurately identify and exploit strategy lineages with deep, sustained evolutionary potential.
% \section{Methodology}

% \subsection{Problem Formulation}
% We formulate the AHD process as maximizing a fitness function $f: \mathcal{X} \rightarrow \mathbb{R}$, where $\mathcal{X}$ is the discrete space of syntactically valid heuristic programs. Unlike RL, which explores in the action space (token-by-token), ES-AHD performs exploration in the parameter (semantic) space.

\section{Methodology: The ES-AHD Framework}

To overcome the limitations of individual-level crossover and mutation in conventional Genetic Algorithms, we formulate the heuristic design process as a continuous-like optimization problem in a high-dimensional semantic space. Our framework, ES-AHD, is directly inspired by the ES paradigm. Let $\mathcal{X}$ denote the discrete space of all syntactically valid heuristic programs, and $f: \mathcal{X} \rightarrow \mathbb{R}$ denote the fitness function representing the algorithmic performance on a target combinatorial optimization problem.
\subsection{Semantic Recombination via LLMs}

In the ES framework, the algorithm maintains a population of $\lambda$ individuals and selects the top $\mu$ performing individuals to update the search center (the mean $m$). Mathematically, the recombination is an arithmetic average:\begin{equation}m^{(g+1)} = \frac{1}{\mu} \sum_{i=1}^{\mu} x_{i:\lambda}^{(g)}\end{equation}where $x_{i:\lambda}^{(g)}$ is the $i$-th best individual at generation $g$.In the context of Automatic Heuristic Design, directly averaging discrete code fragments is mathematically intractable and logically flawed. Instead, we introduce \textit{Semantic Recombination}. Let $\mathcal{P}^{(g)} = \{x_1^{(g)}, x_2^{(g)}, \dots, x_\lambda^{(g)}\}$ be the population of heuristic codes at generation $g$. We select the top $\mu$ individuals to form the elite set $\mathcal{E}^{(g)} = \{x_{1:\lambda}^{(g)}, \dots, x_{\mu:\lambda}^{(g)}\}$.We utilize the advanced contextual reasoning and summarization capabilities of Large Language Models (LLMs) as a semantic aggregation operator, denoted as $\Phi_{\text{reflect}}$. The updated semantic search center, or ``Core Insight'' $\mathcal{I}^{(g)}$, is computed as:\begin{equation}\mathcal{I}^{(g)} = \Phi_{\text{reflect}}\left( \mathcal{E}^{(g)} \right)\end{equation}Here, $\mathcal{I}^{(g)}$ resides in a high-dimensional natural language embedding space. By explicitly abstracting the algorithmic design patterns from the elite set, $\mathcal{I}^{(g)}$ serves as the exact semantic equivalent of the distribution mean $m^{(g)}$ in traditional ES, shifting the search paradigm from point-to-point blind mutation to center-guided directional sampling.

\subsection{Stochastic Covariance Adaptation via Temperature Sampling}
To dynamically address the exploration-exploitation dilemma while avoiding premature convergence, we introduce a stochastic adaptation mechanism for the covariance (Temperature). Instead of a deterministic monotonic decay, the temperature is updated by sampling from a normal distribution, modulated by a constant step-size $\delta \in (0,1)$:
\begin{equation}
    T^{(g)} = \max\left(T_{min}, (1-\delta) \cdot T^{(g-1)} + \delta \cdot \epsilon^{(g)}\right), \quad \epsilon^{(g)} \sim \mathcal{N}(0,1)
    \label{eq:temperature}
\end{equation}
In this formulation, the current temperature maintains a momentum $(1-\delta)$ from the previous generation, while introducing Gaussian noise scaled by the step-size $\delta$. To ensure valid LLM inference and prevent the search variance from collapsing entirely, we explicitly introduce a strictly positive lower bound $T_{min} > 0$. This stochastic random walk in the covariance space allows the framework to primarily shrink the search radius for micro-level refinement, yet retain the critical ability to occasionally sample higher temperatures to escape semantic local optima.

\begin{figure*}[t]
  \centering 
    \includegraphics[width=0.85\textwidth]{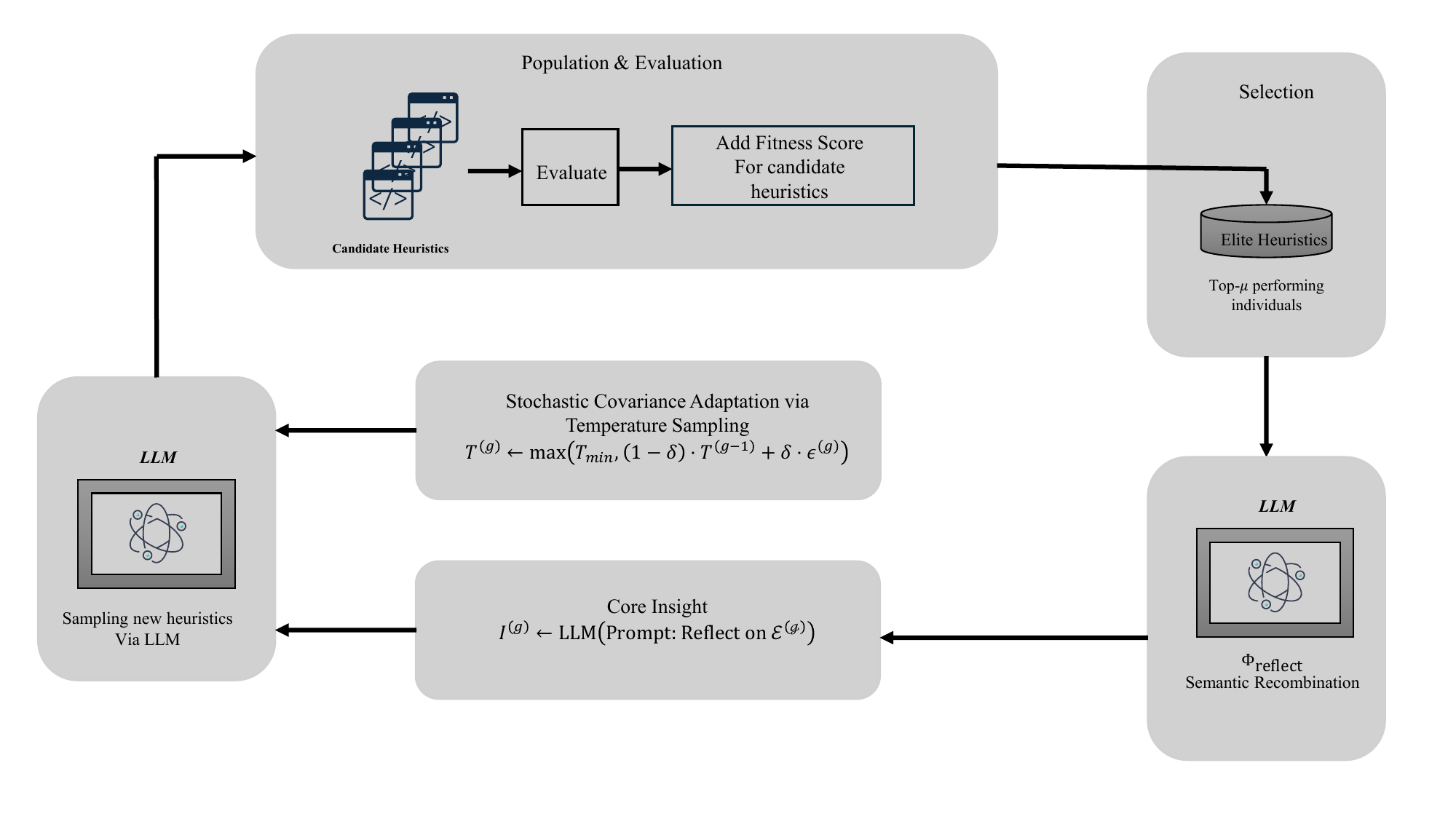}
    % \captionsetup{font={footnotesize}}
  \caption{Overall framework of the proposed ES-AHD.}
  \label{fig:ES-AHD}
\end{figure*}

\subsection{Overall ES-AHD Framework}
The workflow diagram of this paper is shown in Figure~\ref{fig:ES-AHD}. Given a natural language description of the target combinatorial optimization problem $\mathcal{D}$ and a target fitness function $f(\cdot)$ (e.g., the objective value computed on training instances), the algorithm continuously updates a semantic search center to sample high-performance heuristics (Algorithm \ref{algo:es_ahd}). The detailed execution flow is described as follows:

\textbf{1. LLM-driven Initialization:} Instead of relying on manually crafted warm-starts or random syntax trees, ES-AHD utilizes the LLM in a zero-shot manner. Prompted by the problem description $\mathcal{D}$, the LLM independently samples $\lambda$ syntactically valid heuristic programs under the initial temperature $T_0$, forming the initial population $\mathcal{P}^{(0)}$.

\textbf{2. Evaluation and Selection:} In each generation $g$, the algorithm executes all candidate heuristics $x \in \mathcal{P}^{(g)}$ on the target problem to evaluate their fitness $f(x)$. The top $\mu$ performing heuristics are selected to form the elite set $\mathcal{E}^{(g)}$. Simultaneously, the global best heuristic $x^*$ is continuously tracked and updated.

\textbf{3. Semantic Recombination (Center Update):} Unlike traditional ES that computes an arithmetic mean of numerical weights, ES-AHD employs the LLM to perform a semantic reflection over the elite set $\mathcal{E}^{(g)}$. By prompting the LLM to extract the underlying logic and recurring high-performance patterns from the elites, it synthesizes a textual "Core Insight" $I^{(g)}$. This insight serves as the new distribution mean for the current generation.

\textbf{4. Stochastic Covariance Update:} To control the semantic search variance, the algorithm samples a noise variable $\epsilon^{(g)} \sim \mathcal{N}(0, 1)$ and computes the new temperature $T^{(g)}$ using the constant step-size $\delta$. This stochastic update prevents the search process from stagnating in early generations.

\textbf{5. Covariance-Guided Sampling:} Utilizing the updated semantic center $I^{(g)}$ and the stochastically adapted covariance (temperature) $T^{(g)}$, the LLM independently samples $\lambda$ new candidate heuristics. This step populates the next generation $\mathcal{P}^{(g+1)}$ and fundamentally upgrades random, token-level mutation into a center-guided, parameterized semantic perturbation.

\begin{algorithm}[ht]
\caption{ES-AHD}
\label{algo:es_ahd}
\SetAlgoLined
\KwIn{Problem description $\mathcal{D}$, Population size $\lambda$, Elite size $\mu$, Initial Temperature $T_0$, Constant step-size $\delta$, Maximum Generations $G$}
\KwOut{Best heuristic algorithm $x^*$}

$g \leftarrow 0$\;
$T^{(0)} \leftarrow T_0$\;

\tcp{0. LLM-driven Initialization}
$\mathcal{P}^{(0)} \leftarrow \emptyset$\;
\For{$i=1$ \KwTo $\lambda$}{
    Sample $x_i^{(0)} \sim \text{LLM}(\text{Prompt: } \mathcal{D}, \text{Temp: } T^{(0)})$\;
    $\mathcal{P}^{(0)} \leftarrow \mathcal{P}^{(0)} \cup \{x_i^{(0)}\}$\;
}

\While{$g < G$}{
    \tcp{1. Evaluation and Selection}
    \For{each heuristic $x \in \mathcal{P}^{(g)}$}{
        Execute $x$ on training instances to compute fitness $f(x)$\;
    }
    Select elite set $\mathcal{E}^{(g)}$ based on top $\mu$ performers\;
    Update global best $x^* \leftarrow \text{argmax}_{x}(f(x))$\;
    
    \tcp{2. Semantic Recombination (Center $\mu$)}
    $I^{(g)} \leftarrow \text{LLM}(\text{Prompt: Reflect on } \mathcal{E}^{(g)})$\;
    
    \tcp{3. Stochastic Covariance (Temperature) Update}
    Sample $\epsilon^{(g)} \sim \mathcal{N}(0, 1)$\;
    $T^{(g)} \leftarrow \max(T_{min}, (1 - \delta) \cdot T^{(g-1)} + \delta \cdot \epsilon^{(g)})$\;
    
    \tcp{4. Covariance-Guided Sampling}
    $\mathcal{P}^{(g+1)} \leftarrow \emptyset$\;
    \For{$i=1$ \KwTo $\lambda$}{
        Sample $x_i^{(g+1)} \sim \text{LLM}(\text{Prompt: } I^{(g)}, \text{Temp: } T^{(g)})$\;
        $\mathcal{P}^{(g+1)} \leftarrow \mathcal{P}^{(g+1)} \cup \{x_i^{(g+1)}\}$\;
    }
    $g \leftarrow g + 1$\;
}
\Return $x^*$\;
\end{algorithm}

\section{Experiments}

\subsection{Problem Description: Traveling Salesman Problem (TSP)}
The Traveling Salesman Problem~\cite{TSP}  is a fundamental NP-hard problem in combinatorial optimization. Given a set of cities and the distances between each pair, the objective is to find the shortest possible route that visits every city exactly once and returns to the origin city. In our Automatic Heuristic Design (AHD) task, we prompt the Large Language Model (LLM) to design a greedy construction heuristic, specifically a \texttt{select\_next\_node} function. This function iteratively selects the next city to visit based on the current node, the destination node, unvisited nodes, and the distance matrix, thereby constructing a complete TSP tour.

\subsection{Experimental Setup}
To empirically evaluate the effectiveness of the proposed ES-AHD framework, we conduct experiments on TSP instances of different scales.

\begin{itemize}
    \item \textbf{Problem Scale:} We evaluate the algorithms on TSP instances with $N=20, 50, 100$. Testing across this spectrum allows us to comprehensively assess the optimization accuracy, robustness, and scalability of the generated heuristics.
    
    \item \textbf{Evaluation Metric:} We adopt the \textit{Top-4 Average Score} as our primary training performance metric. This score represents the negative total route distance achieved by the top 4 generated heuristic functions on the training set; a higher score indicates a shorter route and thus better performance. For validation, we report the average route distances on the test sets for each respective scale (Val20, Val50, Val100).
    
    \item \textbf{LLM Configuration:} To ensure a fair comparison, all LLM-based AHD methods in our experiments utilize the GLM 4-Flash~\cite{glm2024chatglm} as the underlying heuristics generation engine.
\end{itemize}

\subsection{Baseline Algorithms}
We compare ES-AHD against state-of-the-art LLM-driven AHD baselines:

\begin{enumerate}
    \item \textbf{EoH~\cite{liu2024evolution}:} An evolutionary paradigm that leverages LLMs as crossover and mutation operators to search for heuristics directly in the code space.
    \item \textbf{ReEvo~\cite{ye2024reevo}:} An advanced rewriting-based evolutionary method that focuses on iteratively refining and optimizing existing heuristic code via LLM prompting.
    \item \textbf{FunSearch~\cite{romera2024mathematical}:} A pioneering approach introduced by Google DeepMind that pairs a pre-trained LLM with an automated evaluator to discover novel functions in the program space.
\end{enumerate}

\subsection{Experimental Results and Analysis}
Table \ref{tab:tsp_results} presents the comparative results on the TSP $N=50$ instances.
% % 注意这里的 table* 带有星号
% \begin{table}[t]
% \centering
% \footnotesize
% \caption{Quantitative comparison of the proposed ES-AHD and baseline algorithms. The table reports the top-4 average training scores alongside validation results across problem scales ranging from $N=20$ to $200$.}
% \label{tab:tsp_results}
% \begin{tabular}{lccccc}
% \toprule
% Method & Train (top\_4\_avg\_score)& Val20  & Val50 & Val100 & Val200  \\
% \midrule
% funsearch & -6.364374 & 4.201934 & 6.555002 & 9.021033 & 12.646508 \\
% eoh       & -6.345256 & 4.311166 & 6.696569 & 9.101851 & 12.761167 \\
% reevo     & -6.233072 & 4.182170 & 6.517856 & 8.962170 & \textbf{12.493815} \\
% ES-AHD    & \textbf{-6.208690} & \textbf{4.124913} & \textbf{6.449381} & \textbf{8.911328} & 12.541497 \\
% \bottomrule
% \end{tabular}
% \end{table}
\begin{table}[t]
\centering
% \footnotesize
\caption{Quantitative comparison of the proposed ES-AHD and baseline algorithms. The table reports the top-4 average training scores alongside validation results across problem scales ranging from $N=20$ to $100$.}
\label{tab:tsp_results}
\begin{tabular}{lcccc}
\toprule
Method & Train& Val20  & Val50 & Val100 \\
\midrule
FunSearch~\cite{romera2024mathematical} & -6.364 & 4.202 & 6.555 & 9.021\\
EoH~\cite{liu2024evolution}       & -6.345 & 4.311 & 6.697 & 9.102\\
ReEvo~\cite{ye2024reevo}     & -6.233 & 4.182 & 6.518 & 8.962 \\
ES-AHD    & \textbf{-6.209} & \textbf{4.125} & \textbf{6.449} & \textbf{8.911} \\
\bottomrule
\end{tabular}
\end{table}

% \begin{table}[ht]
%     \centering
%     \caption{Performance Comparison on TSP $N=50$}
    % \label{tab:tsp_results}
    % \begin{tabular}{l|c}
    %     \toprule
    %     \textbf{Algorithm} & \textbf{Top-4 Avg Score} \\
    %     \midrule
    %     FunSearch & -6.3644 \\
    %     EoH & -6.3453 \\
    %     ReEvo & -6.2331 \\
    %     \textbf{ES-AHD (Ours)} & \textbf{-6.2087} \\
    %     \bottomrule
    % \end{tabular}
% \end{table}

As shown in Table \ref{tab:tsp_results}, the proposed ES-AHD framework achieves the best performance among all evaluated methods. ES-AHD obtains a Top-4 Average Training Score of -6.209, outperforming traditional LLM-based evolutionary methods such as EoH (-6.345) and FunSearch (-6.364). More importantly, ES-AHD achieves the best validation results on Val20, Val50 (6.449), and Val100 (8.911), confirming its strong generalization capabilities and the effectiveness of center-guided semantic sampling on these instance sizes. Compared to the highly competitive baseline ReEvo, ES-AHD maintains a distinct advantage in route quality for scales up to 100 nodes. On the largest evaluated scale, ReEvo slightly edges out ES-AHD. This indicates that while ES-AHD excels at discovering highly efficient routing logic for small to medium-scale problems, rewriting-based refinement remains highly resilient on massive instances. Overall, these results confirm the superiority of integrating the Evolution Strategy framework into the semantic search space of AHD to avoid blind search and accelerate the generation of high-quality heuristics.

For the TSPLib benchmark\cite{tsplibbbbb}, ES-AHD demonstrates substantial performance as shown in Table\ref{tab:tsplib_instance_gap}. As a highly authoritative and widely adopted standard dataset in this domain, TSPLIB encompasses a diverse array of real-world and synthetic instances varying in scale (from dozens to hundreds of nodes) and spatial distribution characteristics. Systematic testing on this dataset not only effectively verifies the optimization accuracy and robustness of our algorithm across diverse problem structures, but also ensures a rigorous, fair, and objective performance comparison with state-of-the-art baselines, such as FunSearch, EoH, and ReEvo.

% Requires: \usepackage{booktabs}
\begin{table}[t]
\centering
% \footnotesize % 缩小字号（如果还不行，换成 \scriptsize）
\caption{Results on TSPLib instances}
\label{tab:tsplib_instance_gap}
\begin{tabular}{lcccc}
\toprule
Instance & FunSearch~\cite{romera2024mathematical} & EoH~\cite{liu2024evolution} & ReEvo~\cite{ye2024reevo} & ES-AHD \\
\midrule
bayg29 & 12.727743 & 14.539337 & 13.354037 & \textbf{11.345756} \\
bays29 & 11.464521 & 14.814356 & 14.302805 & \textbf{8.572607} \\
berlin52 & 16.866445 & 17.367538 & 17.352315 & \textbf{13.346826} \\
bier127 & 18.324350 & 20.670233 & 17.854136 & \textbf{14.202483} \\
brazil58 & 15.755070 & 18.224388 & 15.745882 & \textbf{10.864672} \\
brg180 & 398.632479 & 593.888889 & 573.418803 & \textbf{236.794872} \\
burma14 & -98.920470 & -98.926984 & -98.938303 & \textbf{-98.959191} \\
ch130 & 15.084562 & 17.466121 & 15.359798 & \textbf{11.794707} \\
d198 & 23.500085 & 21.149770 & 23.380654 & \textbf{19.434567} \\
eil51 & 9.797103 & 9.591858 & 12.405251 & \textbf{8.017776} \\
eil76 & 11.999104 & 14.126935 & 14.114194 & \textbf{10.543393} \\
gr17 & 8.904876 & 7.965627 & 8.741007 & \textbf{6.666667} \\
gr21 & 15.050486 & 18.652260 & 14.382465 & \textbf{12.350696} \\
gr96 & -98.896594 & -98.916307 & -98.886340 & \textbf{-98.926263} \\
kroB150 & 19.661443 & 20.322299 & 18.929000 & \textbf{18.830394} \\
kroB200 & 20.701744 & 25.341125 & 20.885311 & \textbf{19.322739} \\
kroC100 & 17.597466 & 17.656037 & 17.475751 & \textbf{16.231432} \\
kroD100 & 17.888460 & 18.882770 & 18.126532 & \textbf{17.702952} \\
kroE100 & 16.026355 & 21.408085 & 16.492510 & \textbf{14.395877} \\
lin105 & 20.587496 & 26.108075 & 21.769799 & \textbf{16.233990} \\
pr107 & 14.553533 & 16.269912 & 15.167914 & \textbf{7.376929} \\
pr152 & 21.394295 & 20.440908 & 23.229583 & \textbf{16.738054} \\
pr264 & 26.810131 & 24.458011 & 22.564739 & \textbf{19.068861} \\
pr439 & 23.529199 & 25.997004 & 20.294581 & \textbf{20.231858} \\
si175 & 3.728142 & 3.373507 & 3.368446 & \textbf{2.603510} \\
swiss42 & 13.125164 & 11.377324 & 12.234878 & \textbf{10.480492} \\
u159 & 24.416810 & 22.949131 & 24.949369 & \textbf{18.752916} \\
ulysses16 & -98.733985 & -98.778735 & -98.794378 & \textbf{-98.808491} \\
\bottomrule
\end{tabular}
\end{table}

\section{Conclusion}
In this paper, we introduced ES-AHD, a novel framework that fundamentally integrates the Evolution Strategy paradigm into LLM-driven Automatic Heuristic Design. To overcome the blind search and convergence imbalance inherent in traditional individual-level code mutation, we proposed two core mechanisms. First, Semantic Recombination leverages the LLM to extract center-guided insights from elite individuals, transforming random mutation into highly directional sampling. Second, Stochastic Covariance Adaptation dynamically regulates the semantic search radius via temperature sampling, smoothly balancing micro-level code refinement with the ability to escape local optima. Extensive experiments on TSP instances of varying scales demonstrate that ES-AHD significantly outperforms state-of-the-art baselines, exhibiting superior optimization accuracy and robust generalization. Future work will focus on extending this ES-based framework to more complex combinatorial optimization domains (e.g., vehicle routing and 3D bin packing) and exploring its integration with macro-level tree search strategies to further enhance sampling efficiency in massive heuristic spaces.

\bibliographystyle{ieeetr}
\bibliography{ref}

\end{document}